\documentclass[letter]{ieice}
\usepackage[pdftex]{graphicx,xcolor}
\usepackage[fleqn]{amsmath}
\usepackage{newtxtext}
\usepackage[varg]{newtxmath}

\newcommand{\AmSLaTeX}{%
 $\mathcal A$\lower.4ex\hbox{$\!\mathcal M\!$}$\mathcal S$-\LaTeX}
\def\BibTeX{{\rmfamily B\kern-.05em
 \textsc{i\kern-.025em b}\kern-.08em
  T\kern-.1667em\lower.7ex\hbox{E}\kern-.125emX}}
\makeatletter
\def\tmpcite#1{\@ifundefined{b@#1}{\textbf{?}}{\csname b@#1\endcsname}}%
\makeatother

\field{A}
\vol{98}
\no{1}
\SpecialSection{\LaTeXe\ Class File for the IEICE Transactions}
\title[Adaptive]
      {Adaptive Multi-Agent Continuous Learning System}
\authorlist{%
 \authorentry{Xingyu Qian}{n}{labelA,labelB}\MembershipNumber{}
 \authorentry{Aximu Yuemaier}{n}{labelA,labelB}\MembershipNumber{}
 \authorentry{Longfei Liang}{n}{labelC}\MembershipNumber{}
 \authorentry{Wen-Chi Yang}{n}{labelC}\MembershipNumber{}
 \authorentry[chenxg@mail.sim.ac.cn]{Xiaogang Chen}{n}{labelA}\MembershipNumber{}
 \authorentry{Shunfen Li}{n}{labelA}\MembershipNumber{}
 \authorentry{Weibang Dai}{n}{labelA}\MembershipNumber{}
 \authorentry{Zhitang Song}{n}{labelA}\MembershipNumber{}
}
\affiliate[labelA]{The author is with Chinese Acad Sci, Shanghai Institute of Microsystem and Information Technology, Shanghai 200050, China}
\affiliate[labelB]{The author is with Shanghaitech University, Shanghai 201210, China}
\affiliate[labelC]{The author is with NeuHelium Co., Ltd.}

\begin{document}
\maketitle
\begin{summary}
We propose an adaptive multi-agent clustering recognition system that can be self-supervised driven, based on a temporal sequences continuous learning mechanism with adaptability. The system is designed to use some different functional agents to build up a connection structure to improve adaptability to cope with environmental diverse demands, by predicting the input of the agent to drive the agent to achieve the act of clustering recognition of sequences using the traditional algorithmic approach. Finally, the feasibility experiments of video behavior clustering demonstrate the feasibility of the system to cope with dynamic situations. Our work is placed here\footnote{https://github.com/qian-git/MAMMALS}.
\end{summary}
\begin{keywords}
Continuous Learning, Multi-agent System, Prediction, Adaptability
\end{keywords}

\section{Introduction}
Multi-agent systems are extremely adaptable to deal with complex problems, especially after combining reinforcement learning algorithms, there has been more research progress ~\cite{Hernandez2019Survey,Gronauer2022Multi}.
Independent training approaches such as IQL can lead to difficult learning convergence because of the shared dynamic instability environment factor, while centralized execution, such as VDN integrates the value function of each intelligence, and the performance can be improved even more by ignoring individual characteristics.
And most of the multi-agent systems use the centralized training decentralized execution (CTED) approach, such as MADDPG~\cite{Multi2017} and QMIX~\cite{2018QMIX}.
However, for the centralized training approach, the multi-agent system always uses the system goal to evaluate the computational tasks and training processes of the agents, and even in some cases, the learning is performed in a simulator or laboratory ~\cite{Foerster2018}, in which additional state information is available and the agents can communicate freely through global data sharing.
In contrast, the decentralized execution approach uses a strategy of decentralizing the computational tasks, where each agent performs its learning process with local data and the environment.
And the decentralized approach also reduces the complexity of interaction with the increased number of agents and avoids partial observability challenges.
Therefore, it is always necessary to find an effective decentralized execution strategy for both the execution process and the training process.
\section{Motivation}
Luczak~\cite{luczak2022} found that a single neuron is not only performing excitation, but it also predicts the future, through algorithmic validation of prediction systems and experimental studies of biological neurons. It was shown that prediction mechanisms may be an important processing element in neurons and that internal prediction models exist in neurons as a key component of the brain's learning mechanisms.
In addition, it has also been hypothesized by related researchers that the brain may perform learning behaviors based on the principle of predictive coding ~\cite{moshe2007the,clark2013what,buz2019the}.
Inspired by this, we propose the adaptive multi-agent clustering recognition system with driving an agent by predicting its input.
The system is designed to meet the diverse demand by combining several different functional agents in a connected structure.
Unlike artificial neural networks, we use the traditional algorithm scheme to design decentralized multi-agent systems oriented to clustering learning tasks, without uniform evaluation standards for the execution state of the agent for centralized management.
To avoid excessive data interference, we choose to simplify the processing behavior by using temporal sequences input, and the real-time change characteristic of temporal sequences indicates the continuous process of stimulus, while the processing logic of a single agent is simplified to the clustering recognition and prediction behavior, taken the clustering results on the time series as the output.
A system formed by the combined connection of multiple agents can theoretically achieve clustered cognition of complex computational behaviors through the prediction-driven mechanism of each agent.
\section{ Clustering Recognition System}
A single agent learns the regularity of the received complete temporal sequences using clustering algorithms, identifies and predicts the received data for every input using the clustering results, and outputs the clustering and prediction results.
Each agent has four data interfaces, which are sequence-receiving and encoding-sending interfaces related to temporal sequence, and prediction-sending and feedback-receiving interfaces related to prediction.
In the system hierarchy, we design a multi-layer structure to establish connections between agents, and the upper and lower layers of agents use a many-to-one approach to perform data interaction, the closer to the top, the fewer agents there are.
The agent at the bottom level acts as the input of the system, and the agent at the top level is the output with only one agent, the system structure is similar to a pyramid form.
The encoding-sending interface of the agent at the lower level connects to the sequence-receiving interface of the agent at the higher level to aggregate the clustering behaviors of the agents at the lower level to assemble a new temporal sequence; the prediction-sending interface at the higher level connects to the feedback-receiving interface of each agent at the corresponding lower level to propagate the prediction results as feedback information.
The predicted output of an agent is sent in the opposite direction of the sequence. 
The clustering results are sent to higher-level agents as codes, while the prediction results are sent to lower-level agents.
The feedback received by the agents when comparing the temporal sequence for recognition can help the clustering result selection in the comparison process and improve the recognition performance.
\subsection{self-supervised driving}
For the internal execution part of the agents, similar to what was proposed in ~\cite{luczak2022} where a single neural can generate predictive behavior for the future based on external excitations, we designed to enable each agent to implement a clustering learning behavior of the input temporal sequence under the condition that external incentives are simulated using temporal sequence, and gradually develop clustering results for different temporal sequences in the learning process.
Using predictive behavior to drive the agent to perform the clustering learning process is designed for the agent to always predict the input data more accurately.
To go to the correct prediction as much as possible, the agent must identify the received individual time series for more accurate classification and use the matched clustered data for prediction.
And verification of the agent's prediction evaluation only requires the next input to validate the prediction.
The learning of the system is a continuous process, and the agents that receive temporal sequences in the system can perform the computation process, while the agents that choose to disconnect or not established connections do not have valid data input, so they are not driven by predictions and in a "standby" state.
\subsection{generic}
The system uses temporal sequences as the usage and data exchange method for all agents, and if events can be converted into temporal sequences coding, they can all be learned and predicted under this system, such as natural language processing and video behavior clustering recognition.
Corresponding to different application scenarios, the way of combining connection relations between agents can be specially designed.
For example, for high-density information with spatio-temporal characteristics such as video images, multiple agents dealing with neighboring regions jointly search for an agent to establish connections, and the higher agent receives the clustering recognition results of multiple agents at the lower level to achieve a summary of spatio-temporal characteristics in a larger range, thus establishing a multi-layer structure to cover the entire picture range.
\subsection{decentralized parallel learning}
The number of agents used in a system structure is not only related to the corresponding connections between the two layers but also the number of layers of the system design, with the number of agents required increases with the number of layers of the system.
The centralized training and system evaluation learning requires the information of all agents to evaluate the learning performance, and the size of the agents will undoubtedly lead to problems such as the curse of dimensionality.
On the one hand, In our system, the agent system does not rely on a centralized evaluation approach, but rather the agents perform self-supervised verification to judge the prediction based on the input data.
This makes the agents become a series of closed subsystems, except for the necessary data exchange.
On the other hand, the temporal sequences received by each layer of agents will only stay in the agents of this layer and will not be propagated further, which makes the propagation of data streams in the system proceed in a relay manner and makes the data exchange more orderly.
The multilayer structure will converge the learning and recognition effects of each layer of agents, and the temporal sequence received by the top agent represents the aggregated feature amount of the changing pattern jointly summarized by all other agents, while the system's task objective is achieved by the clustering recognition results of the top agent.
\subsection{co-convergence}
Multi-agent systems inherently suffer from the complexity problems of multi-agent, and often the computational workload grows exponentially with the number of agents, while also requiring agents to be highly adaptable in complex environments.
Each agent performs a new clustering learning behavior to improve the prediction effect, and the higher agent learns inductively the clustering behavior of the previous agent. If the clustering results of the lower level become divergent, it will accordingly lead to the failure of all the agents in the higher level to converge.
We use a many-to-one approach to design the connection structure of the agents while passing the prediction of the higher level to the lower level as feedback information, to converge the learning behaviors of each agent.
Since the lower-level agent is transmitting each clustering result to the higher level, for the higher-level agent, its received temporal sequence can also be understood as the change process of the clustering behavior of the previous level agent.
The predicted value of an agent for the input can then be understood as the change in behavior based on the future execution of the lower-layer agent.
Therefore, the lower layer agents receive this prediction to know in advance the future choices that will be made, which allows a smooth convergence to take place quickly with feedback information in case of oscillations in the recognition process.

Take the sequence for example: “ABCABDADD...”, agents A and B can both perform recognition of three-character segments.
Agent A has coded three recognizable results for "ABC", "ABD", and "ADD" to 1, 2, and 3 respectively.
When this same sequence is entered again because there is a similar "AB", the recognition process of agent A may become "1111112133...", while agent B can easily get the recognition pattern of "123" from A's final choice.
This makes it possible for agent A to select "111222333..." directly from the prediction of agent B after recognizing the "ABC" fragment. 
In the sequence example, the agent in the first level looks for the data change pattern from the relatively short-range sequence, and the agent in the later level summarizes the pattern of the previous level, thus realizing the recognition of the long-range pattern.
And agent A uses the prediction results of B to get stable identification and prediction results in advance, the data that A continues to pass to B will also become stable in advance, and this can also help to converge the effect of B clustering.

In addition, the many-to-one approach makes it possible that even if one agent in the lower level consistently gives incorrect results, the correct identification of most agents can make the higher-level agent still perform the correct clustering behavior, and the oscillations caused by the lower-level agent do not affect the higher layer, while the prediction results also promote the adjustment of the incorrectly clustered agents and finally achieve a smooth convergence of the overall system.
\section{experiment and analysis}
In the end, we conducted a feasibility test experiment on the system, and we chose vehicle behavior recognition in video analysis as our experimental scene, with data collected from the Rome Street View traffic intersection ~\cite{MAMMALS}, where different types of vehicle driving styles exist.
A hierarchical connection structure is designed, where the last layer uses 1 agent to cover the whole screen area as the final system recognition result.
Each agent is a summary and prediction of the image change pattern of the region it covers, and through this bottom-up layer-by-layer learning approach, the overall goal for video behavior recognition can be obtained by the clustering result of the agent in the highest layer.
For the clustering learning behavior of the agent, we use the longest common subsequence that conducts the public similarity comparison as the comparison algorithm of the agent clustering learning, and after each clustering of the complete temporal sequence, the final matching clustering sequence is also maintained in time using this time series to achieve the effect of continuous updating of the clustering results.

\begin{figure}[ht]
  \centering
	\includegraphics[width=1\linewidth]{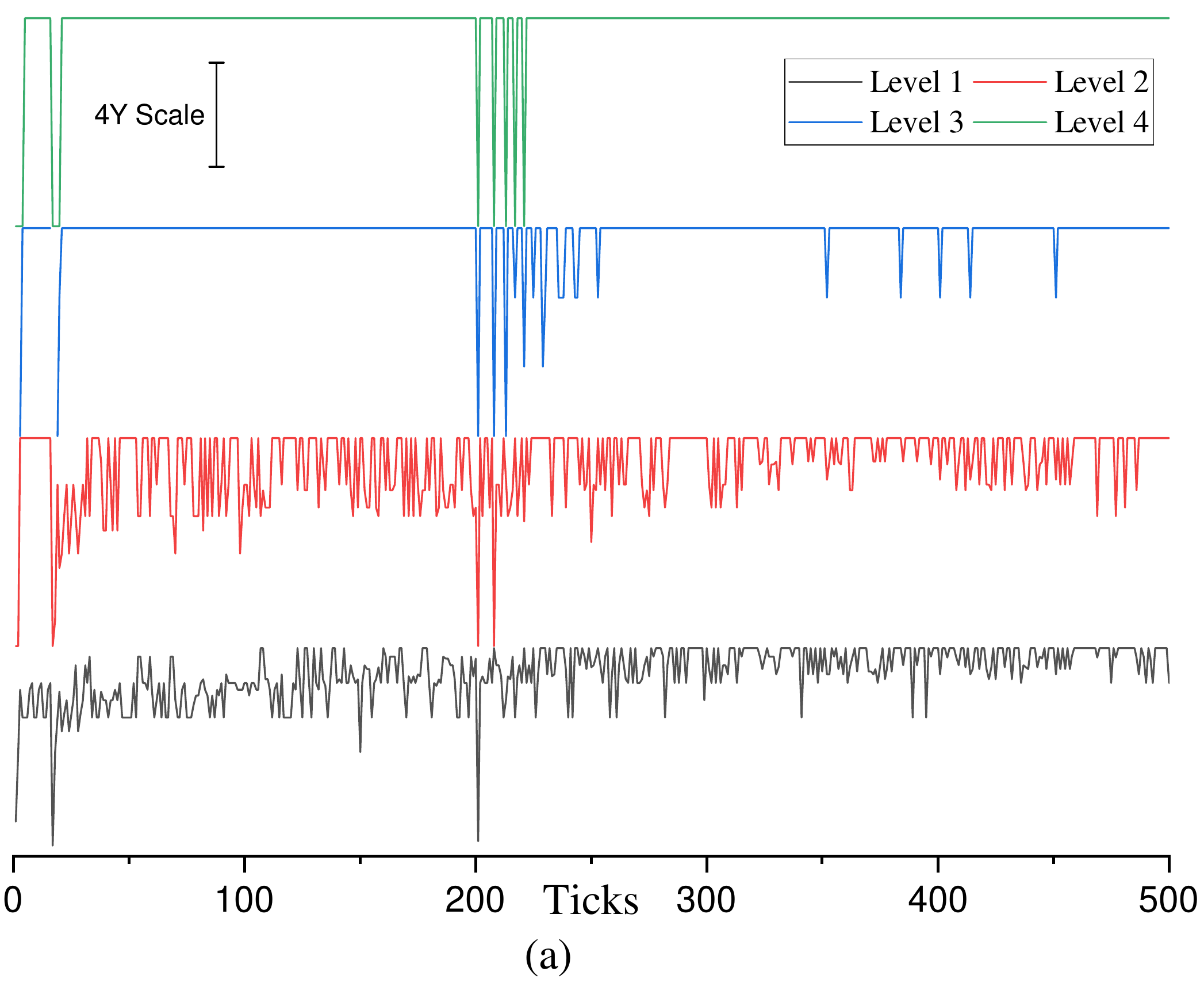}
	\\
	\includegraphics[width=1\linewidth]{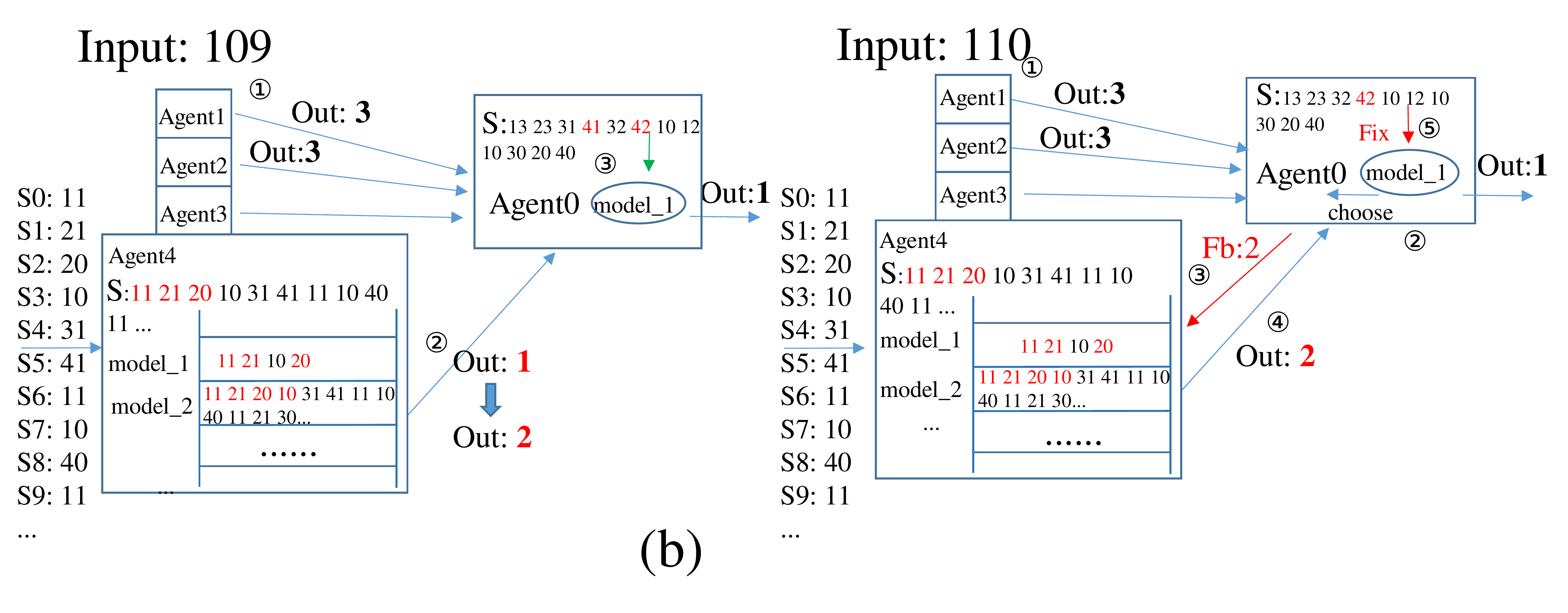}
	\\
	\includegraphics[width=0.49\linewidth]{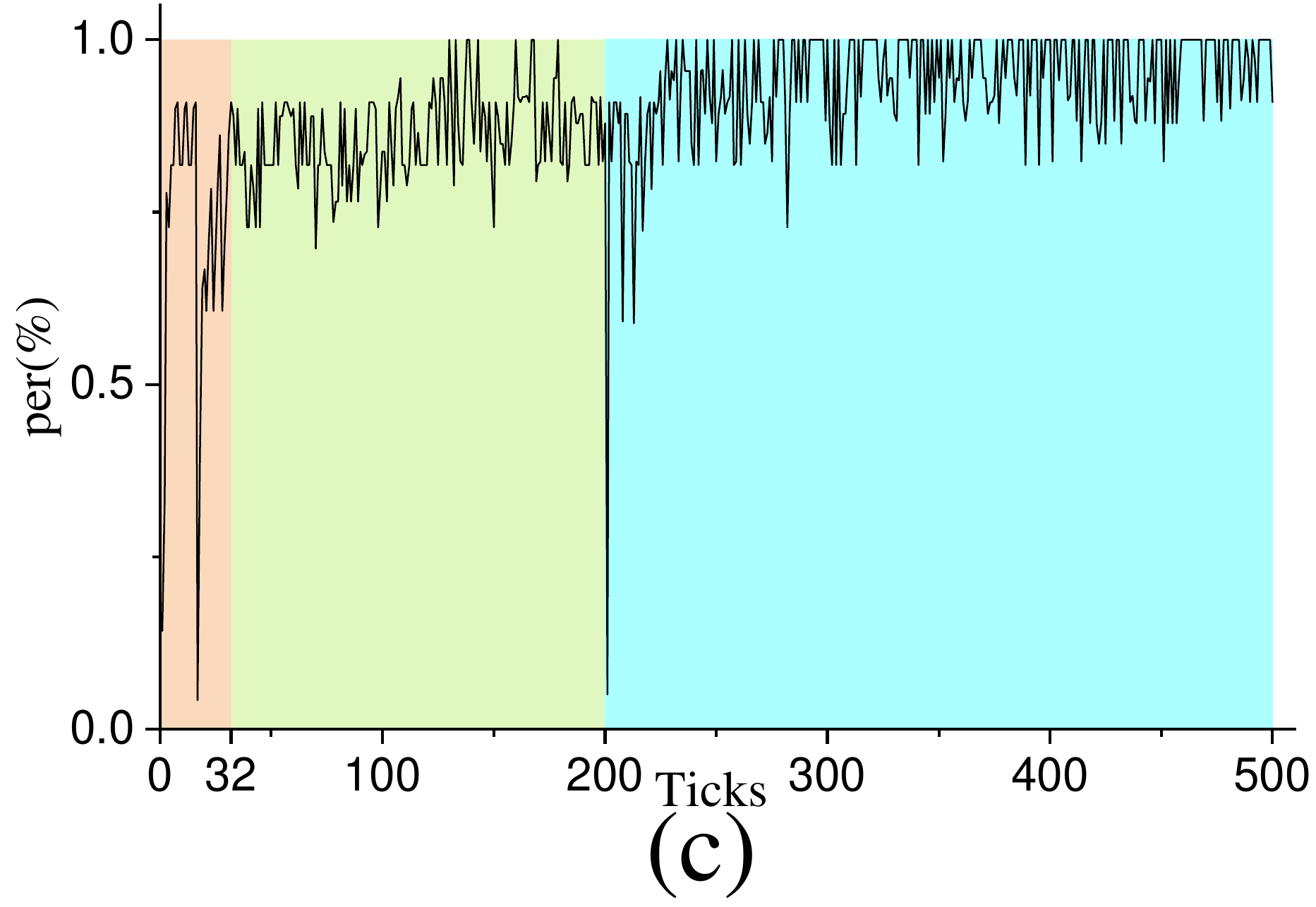}
	\
  \includegraphics[width=0.49\linewidth]{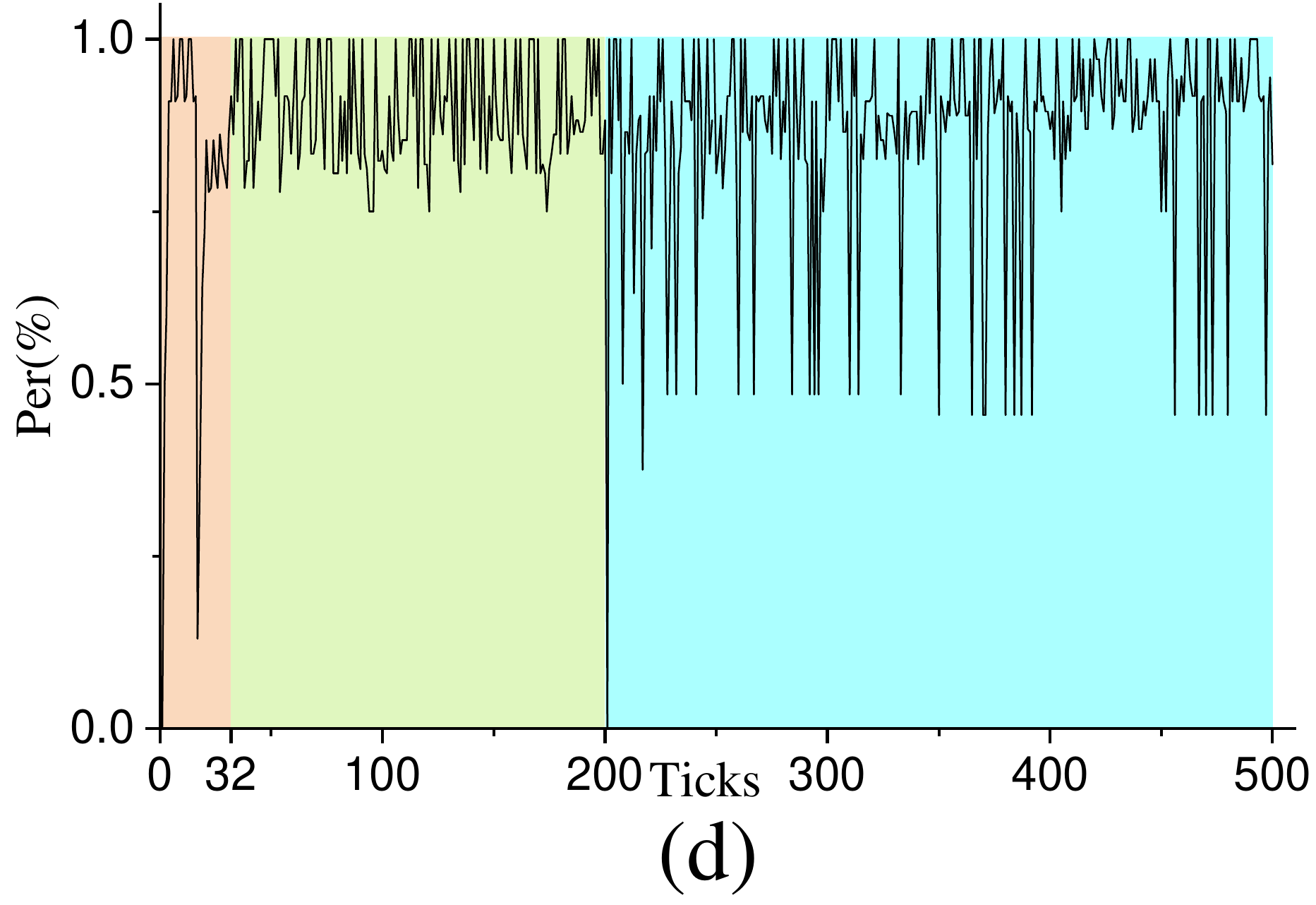}
  \caption{clustering resultm}
  \label{pic1}
\end{figure}

The overall experiment continues for a total of 500 times which can be divided into three types of vehicle video input, two of which are randomly sorted throughout, and the third middle type is randomly inserted after the 200th time, the experimental results are shown in Figure \ref{pic1}.
Figure~\ref{pic1}. (a) shows Y-axis offset stacking of the correct clustering results of the four layers of agents. 
Figure~\ref{pic1}. (b) shows the interaction between several agents in the first two layers during two consecutive inputs in the experiment.
The horizontal coordinates indicate the number of test inputs, and the vertical coordinates indicate the proportion of agents that clustered correctly.
The first 32 times (orange part) are the system start-up phase, where two types of behavioral sequential inputs are performed to help initial model building, the green part is a random test of two types of behaviors, and the blue part starts from the 200th time, where a third type is gradually added during the random test until the end of the test.

Figure ~\ref{pic1}. (c) shows the correct clustering rate of the whole system agent, and it can be found that in the initial stage of testing and for a period of 200 tests, the absence of the model of interest leads to poor clustering results.
However, as the testing continues, there is a continuous upward trend, which indicates that the system can continuously improve the learning effect.
Figure ~\ref{pic1}. (a) can find that as the number of layers increases, the clustering behavior effect of the agents converges more flatly.
This is because by connecting multiple neighboring agents to the same agent, a larger range of spatio-temporal characteristics learning can be obtained, where a few regions are difficult to produce correct results after the first level agent can still summarize the changing pattern from the results of most agents.
Figure ~\ref{pic1}(b) shows this specific inter-agent cooperative behavior.
This is the case of agent~1-4 in the first layer and agent~0 in the second layer in the 109th and 110th tests, where the two inputs are the same because the overall change process is similar, and the difference is that there is a feedback behavior in the 110th test.
There is a spatial-temporal correlation in the behavior change, so agent~4 data came later than agent~1 and agent~2.
In the 109th test, because there is no feedback from Agent~0, Agent~4 has an oscillation in recognition results, and this change is also passed to Agent~0.
However, Agent~0 is less affected because of the correct results of the other three agents. 
In the 110th test, the feedback from Agent~0 makes Agent~4 make the correct choice in advance, and the output no longer oscillates, which also makes the execution activity of Agent0 stable.

And Figure~\ref{pic1}(d) is the overall clustering correctness result for the same experimental conditions with the predicted feedback behavior between agents removed.
Compared with Figure~\ref{pic1}(c), the clustering effect of the system does not change much as the test continues, and the clustering effect becomes worse after adding new change cases and can no longer be converged.
This indicates that the feedback behavior between agents does improve the learning-thinning effect of the system.
\section{summary}
Facing the dynamically changing environment, it has been a challenge for multi-agent systems to design the system structure and task objectives to improve the response to change adaptability.
Facing the diverse needs of environmental changes, we propose an adaptive multi-agent clustering recognition system that uses time series to achieve recognition by clustering predictive behaviors.
The agent internally completes the clustering prediction activity on the input, and the system achieves the diversity requirement by designing the connection structure with different combinations of agents with different functions.
Finally, the feasibility test experiments are conducted for video behavior recognition scenarios, and the experiments yield correct recognition results, while the synergistic effect between agents and the form of connection combinations can achieve convergence in the results, which shows the feasibility of the system to meet the diversity requirements.
However, there are still many factors that can be explored.
For example, in the selection of the number of layers of the system, we also conducted a test session for 5-layer and even 6-layer structures designed.
The increase in the number of agents at the bottom layer allows the system to obtain finer classification results but also makes the convergence rate of the correct clustering rate of all agents slower.
Another example is the design of the connection structure of the agents for self-organization to establish connection relations, so that the multi-layer structure can be implemented more flexibly according to the state of neighboring agents, etc.
Finally, the feasibility test experiments on video behavior recognition scenarios are conducted to verify the synergistic effect between agents and the convergence of results achieved in the face of new changes, which shows the feasibility of the system to meet the diversity requirements.

\section*{Acknowledgments}
This project is supported by the Strategic Priority Research Program of the Chinese Academy of Sciences (XDB44010200), National Natural Science Foundation of China (92164302, 61874129, 91964204, 61904186, 61904189, 61874178), Science and Technology Council of Shanghai (17DZ2291300, 19JC1416801, 2050112300), the Youth Innovation Promotion Association CAS under Grant 2022233. It was also supported by Shanghai R$\&$D and Transformation Functional Platform Project(grant agreement number 17DZ2260900): A Functional Platform of Neuromorphic Chips and Intelligent Systems On-chip

\bibliographystyle{ieicetr}
\bibliography{reference}

\end{document}